\documentclass[runningheads]{llncs}
\usepackage{graphicx}
\usepackage{tikz}
\usepackage{comment}
\usepackage{amsmath,amssymb} % define this before the line numbering.
\usepackage{color}
\usepackage{multirow}
\usepackage{booktabs}
\usepackage{subcaption}

\usepackage[accsupp]{axessibility}  % Improves PDF readability for those with disabilities.

\usepackage[width=122mm,left=12mm,paperwidth=146mm,height=193mm,top=12mm,paperheight=217mm]{geometry}

\def \ie{\textit{i.e.}}
\def \eg{\textit{e.g.}}
\def \hfillx {\hspace*{-\textwidth} \hfill}

\begin{document}
% \linenumbers
\pagestyle{headings}
\mainmatter
\def\ECCVSubNumber{xxxx}  % Insert your submission number here

\title{Structured Context Transformer for Generic Event Boundary Detection} % Replace with your title

\begin{comment}
\titlerunning{ECCV-22 submission ID \ECCVSubNumber} 
\authorrunning{ECCV-22 submission ID \ECCVSubNumber} 
\author{Anonymous ECCV submission}
\institute{Paper ID \ECCVSubNumber}
\end{comment}
%******************

% \begin{comment}
\titlerunning{SC-Transformer}
\author{Congcong Li, Xinyao Wang, Dexiang Hong, Yufei Wang, Libo Zhang, Tiejian Luo, Longyin Wen}
\institute{
{University of Chinese Academy of Sciences, Beijing, China } \\
{ByteDance Inc., Mountain View, USA}\\
{Institute of Software Chinese Academy of Sciences, Beijing, China}\\
}
\authorrunning{Congcong et al.}
% \end{comment}
%******************
\maketitle

\begin{abstract}Generic Event Boundary Detection (GEBD) aims to detect moments where humans naturally perceive as event boundaries. In this paper, we present Structured Context Transformer (or SC-Transformer) to solve the GEBD task, which can be trained in an end-to-end fashion. Specifically, we use the backbone convolutional neural network (CNN) to extract the features of each video frame. To capture temporal context information of each frame, we design the structure context transformer (SC-Transformer) by re-partitioning input frame sequence. Note that, the overall computation complexity of SC-Transformer is linear to the video length. After that, the group similarities are computed to capture the differences between frames. Then, a lightweight fully convolutional network is used to determine the event boundaries based on the grouped similarity maps. To remedy the ambiguities of boundary annotations, the Gaussian kernel is adopted to preprocess the ground-truth event boundaries to further boost the accuracy. Extensive experiments conducted on the challenging Kinetics-GEBD and TAPOS datasets demonstrate the effectiveness of the proposed method compared to the state-of-the-art methods.
\end{abstract}

\section{Introduction}
Video has accounted for a large part of human's life in recent years. Aided by the rapid developments of hardware, video understanding has witnessed an explosion of new designed architectures \cite{DBLP:conf/icml/3D-Convolutional,DBLP:conf/iccv/Learning-Spatiotemporal,DBLP:conf/iccv/SlowFast,DBLP:conf/cvpr/Two-Stream,DBLP:journals/corr/abs-2102-00719,DBLP:journals/corr/abs-2106-13230} and datasets \cite{DBLP:journals/corr/Kinetics,DBLP:journals/corr/UCF101,DBLP:conf/cvpr/TAPOS,DBLP:conf/iccv/HMDB,DBLP:conf/cvpr/PerazziPMGGS16}. The cognitive science \cite{tversky2013event} suggests that humans naturally divide video into meaningful units. To enable machines to develop such ability, Generic Event Boundary Detection \cite{DBLP:journals/corr/GEBD} (GEBD) is recently proposed which aims at localizing the moments where humans naturally perceive event boundaries.

\begin{figure}[t]
    \centering
    \includegraphics[width=\textwidth]{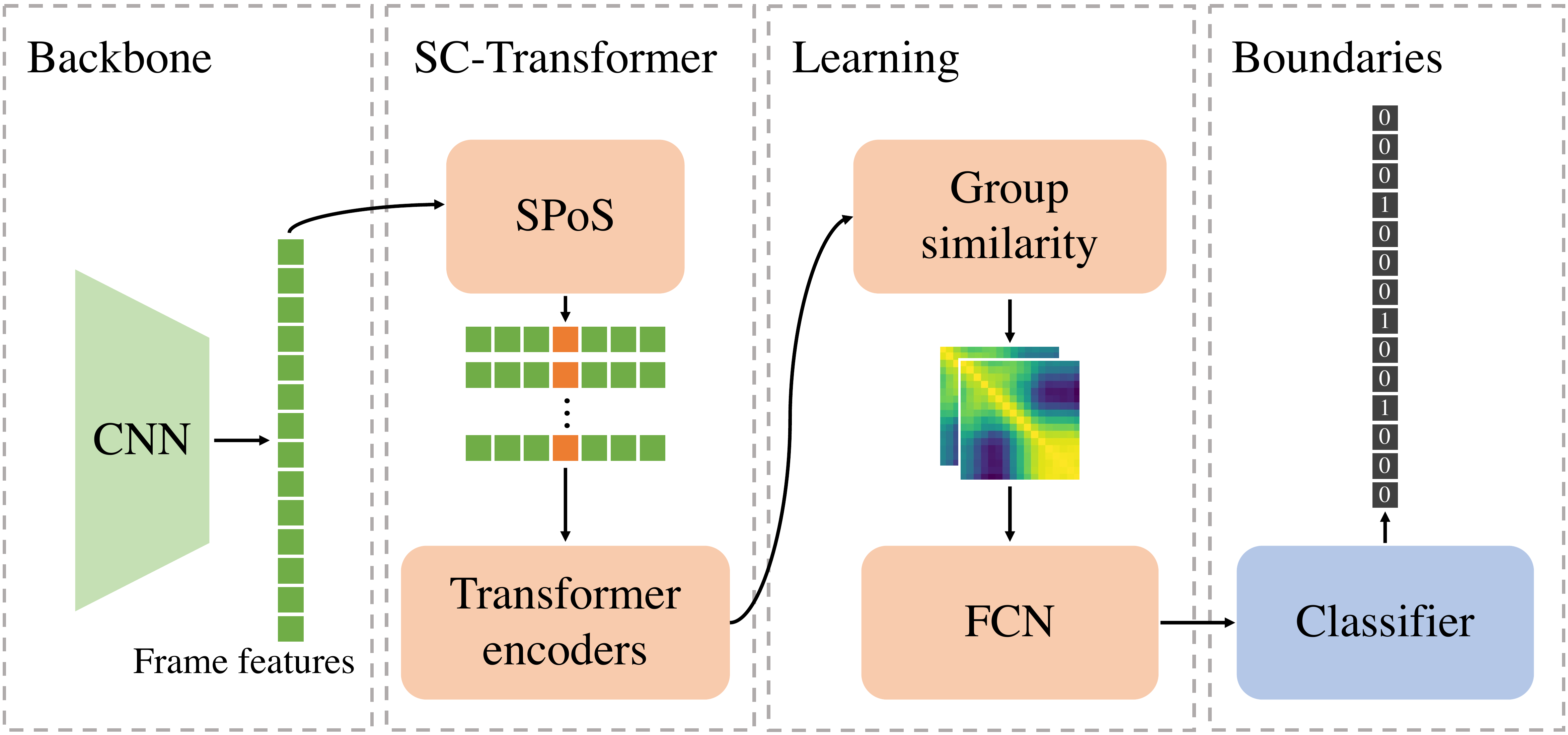}
    \caption{Overview architecture of the proposed method. The proposed method can predict all boundaries of video sequences in a single forward pass with high efficiency. We use a CNN backbone to extract the 2D features of each video frame. These features are then pooled into the vectors and converted into a sequence. The structured partition of sequence (SPoS) mechanism is employed to
    re-partition input frame sequence and provides \textbf{structured context} for each candidate frame. Based on this structured context, the transformer encoder blocks are used to learn the high level representations of each local sequence, which have linear computational complexity with respect to the video length and enable feature sharing. After that, we compute the group similarities to encode frame differences and use a lightweight fully convolutional network (FCN) is predict event boundaries based on the computed 2D grouped similarity maps.}
    \label{fig:framework}
\end{figure}

Event boundaries in the GEBD task are taxonomy-free in nature and can be seen as a new attempt to interconnect human perception mechanisms to video understanding. Annotators are required to localize boundaries at ``one level deeper'' granularity compared to the video-level event. To remedy the ambiguities of event boundaries based on human perception, five different annotators are employed for each video to label the boundaries based on predefined principles. These characteristics differentiate GEBD from the previous video localization tasks \cite{DBLP:journals/access/XiaZ20} by several high-level causes, for example, 1)  Change of Subject, \ie, new subject appears or old subject disappears, 2) Change of Action, \ie, an old action ends, or a new action starts, 3) Change in Environment,\ie, significant changes in color or brightness of the environment, 4) Change of Object of Interaction, \ie, the subject starts to interact with a new object or finishes with an old object. The aforementioned factors make GEBD to be a more challenging task compared to video localization.

Solving GEBD task is not trivial since detecting event boundaries highly rely on temporal context information. Existing methods tackle this problem by processing each frame individually \cite{DBLP:journals/corr/GEBD,DBLP:journals/corr/progressive,DBLP:journals/corr/abs-2107-00239} or computing the global self-similarity matrix and using extra parsing algorithm to find boundary patterns based on the self-similarity matrix \cite{DBLP:journals/corr/gebd-UBoCo,DBLP:journals/corr/gebd-contrastive-learning}. The methods in the first category introduce substantial redundant computations of adjacent frames in a video sequence when predicting boundaries and have to solve the class imbalance issue of event boundaries. The methods in the second category have quadratic computation complexity regarding to the length of input videos due to computation of self-attention globally and the usage of the extra parsing algorithm to predict boundaries.

To that end, we proposed an end-to-end method to predict all boundaries of video sequences in a single forward pass of the network with high efficiency. The overall architecture of the proposed method is shown in Figure \ref{fig:framework}. Specifically, the Structured Context Transformer (SC-Transformer) is designed for GEBD based on the designed structured partition of sequence (SPoS) mechanism, which has linear computational complexity with respect to input video length and enables feature sharing by design. Structured partition of sequence (SPoS) mechanism brings the local feature sequences for each frame in a one-to-one manner, which is termed as \textbf{structured context}. We also find that 1D CNNs actually make the candidate frames attend to adjacent frames in a Gaussian distribution manner \cite{DBLP:conf/nips/LuoLUZ16}, which is not optimal for boundary detection as adjacent frames are equally important. Our proposed SC-Transformer can learn a high level representation for each frame within its structured context which is critical for boundary detection. After that, we use the group similarity to exploit discriminative features to encode the differences between frames. The concept of groups as a dimension for model design has been widely studied, including Group Convolutions \cite{DBLP:conf/nips/alexnet,DBLP:conf/cvpr/ResNeXt}, Group Normalization \cite{DBLP:conf/eccv/GN}, Multi-head self attention \cite{DBLP:conf/nips/att_is_all_you_need}, \textit{etc}. However, to the best of our knowledge, there is still no study on the grouped similarity learning. Previous methods \cite{DBLP:journals/corr/gebd-UBoCo,DBLP:journals/corr/gebd-contrastive-learning,DBLP:journals/corr/progressive} actually compute similarity matrix on one dimension group. Our proposed group similarity allows the network to learn a varied set of similarities and we find it is effective for GEBD. Following the group similarity, a lightweight fully convolutional network \cite{DBLP:conf/cvpr/FCN} (FCN) is used to predict event boundaries. Note that, to speed up the training phase, the Gaussian kernel is used to preprocess the ground-truth event boundaries. Extensive experiments conducted on two challenging Kinetics-GEBD and TAPOS datasets demonstrate the effectiveness of the proposed method compared to the state-of-the-art methods. Specifically, compared to DDM-Net \cite{DBLP:journals/corr/progressive}, our method produces 1.3\% absolute improvement. Meanwhile, compared to PC \cite{DBLP:journals/corr/GEBD}, our method achieves 15.2\% absolute improvement with $5.7\times$ faster running speed. We also conduct several ablation studies to analyze the effectiveness of different components in the proposed method. We hope the proposed method can inspire future work.

The main contributions of this paper are summarized as follows. (1) We propose the structured context transformer for GEBD, which can be trained in an end-to-end fashion. (2) To capture differences between frames, we compute the group similarities to exploit the discriminative features to encode the differences between frames and use a lightweight FCN to predict the event boundaries. (3) Several experiments conducted on two challenging Kinetics-GEBD and TAPOS datasets demonstrate the effectiveness of the proposed method compared to the state-of-the-art methods.

\section{Related Works}
\noindent\textbf{Generic Event Boundary Detection (GEBD).} The goal of GEBD \cite{DBLP:journals/corr/GEBD} is to localize the taxonomy-free event boundaries that break a long event into several short temporal segments. Different from TAL, GEBD only requires to predict the boundaries of each continuous segments. The current methods\cite{DBLP:journals/corr/gebd-contrastive-learning,DBLP:journals/corr/abs-2107-00239,DBLP:journals/corr/abs-2106-10090} all follow the similar fashion in \cite{DBLP:journals/corr/GEBD}, which takes a fixed length of video frames before and after the candidate frame as input, and separately determines whether each candidate frame is the event boundary or not. Kang \textit{et al.} \cite{DBLP:journals/corr/gebd-contrastive-learning} propose to use the temporal self-similarity matrix (TSM) as the intermediate representation and use the popular contrastive learning method to exploit the discriminative features for better performance. Hong \textit{et al.} \cite{DBLP:journals/corr/abs-2107-00239} use the cascade classification heads and dynamic sampling strategy to boost both recall and precision. Rai \textit{et al.} \cite{DBLP:journals/corr/abs-2106-10090} attempt to learn the spatiotemporal features using a two stream inflated 3D convolutions architecture.

\noindent\textbf{Temporal Action Localization (TAL).} TAL aims to localize the action segments from untrimmed videos. More specifically, for each action segment, the goal is to detect the start point, the end point and the action class it belongs to. Most approaches could be categorised into two groups, A two-stage method\cite{richard2016temporal,ni2016progressively,caba2017scc,zhao2017temporal,Chao_2018_CVPR} and a single-stage method\cite{lea2017temporal,lin2017single,alwassel2018action,long2019gaussian,Yuan_2017_CVPR,ma2016learning,Yuan_2017_CVPR,zhao2020bottom}. In a two-stage method setting, the first stage generates action segment proposals. The actionness and the type of action for each proposal are then determined by the second stage, along with some post-processing methods such as grouping \cite{zhao2017temporal} and Non-maximum Suppression (NMS) \cite{DBLP:conf/iccv/BMN} to eliminate redundant proposals. For one-stage methods, the classification is performed on the pre-defined anchors\cite{lin2017single,long2019gaussian} or video frames\cite{ma2016learning,Yuan_2017_CVPR}. Even though TAL task has some similarity to GEBD task, there is no straightforward way to directly apply these methods on the GEBD dataset. Since GEBD requires event boundaries to be taxonomy-free and continuous, which is different from the TAL settings.

\noindent\textbf{Transformers.} Transformer \cite{DBLP:conf/nips/att_is_all_you_need} is a prominent deep learning model that has achieved superior performance in various fields, such as natural language processing (NLP) and computer vision (CV). Despite it's success, the computational complexity of its self-attention is quadratic to image size and hard to applied to high-resolution images. To address this issue, \cite{DBLP:conf/iccv/swin-transformer} proposes a hierarchical Transformer whose representation is computed with shifted windows and has linear computational complexity with respect to image size. In this paper, we show that these Transformer variants are not suitable for GEBD.

\section{Method}
\label{sec:method}
% Generic Event Boundary Detection \cite{DBLP:journals/corr/GEBD} (GEBD) aims at localizing the moments where humans naturally perceive event boundaries without the need of a predefined target event taxonomy.
The existing methods \cite{DBLP:journals/corr/GEBD,DBLP:journals/corr/progressive,DBLP:journals/corr/abs-2107-00239} formulates the GEBD task as binary classification, which predict the boundary label of each frame by considering the temporal context information. However, it is inefficient because the redundant computation is conducted while generating the representations of consecutive frames. To remedy this, we propose an end-to-end efficient and straightforward method for GEBD, which regards each video clip as a whole. Specifically, given a video clip of arbitrary length, we first use conventional CNN backbone to extract the 2D feature representation for each frame and get the frame sequence, \ie, $V = \{ I_t\}_{t=1}^T$, where $I_t \in \mathbb{R}^C$ and $T$ is the length of the video clip. Then the structured partition of sequence (SPoS) mechanism is employed to re-partition input frame sequence $\{ I_t\}_{t=1}^T$ and provide \textbf{structured context} for each candidate frame. The Transformer encoder blocks \cite{DBLP:conf/nips/att_is_all_you_need} are then used to learn the high level representation of each local sequence. After that, we compute the group similarities to capture temporal changes and a following lightweight fully convolutional network \cite{DBLP:conf/cvpr/FCN} (FCN) is used to recognize different patterns of the grouped 2D similarity maps. We will introduce the details of each module in the following sections. The overall architecture of proposed method is presented in Figure \ref{fig:framework}.

\subsection{Structured Context Transformer}
\label{sec:sc_transformer}
The existence of an event boundary in a video clip implies that there is a visual content change at that point, thus it is very difficult to infer the boundary from one single frame. As a result, the key clue for event boundary detection is to localize changes in the temporal domain. Modeling in temporal domain has long been explored by different approaches, including LSTM \cite{hochreiter1997lstm}, Transformer \cite{DBLP:conf/nips/att_is_all_you_need}, 3D Convolutional Neural Network\cite{DBLP:conf/iccv/Learning-Spatiotemporal}, \textit{etc}. Transformer \cite{DBLP:conf/nips/att_is_all_you_need} has recently demonstrated promising results on both natural language processing (NLP) tasks and computer vision tasks. Despite its success, it is difficult to apply Transformer directly to the GEBD task due to its quadratic computational complexity of self-attention. The computation cost and memory consumption increase dramatically as the length of video increases. Previous methods \cite{DBLP:journals/corr/GEBD,DBLP:journals/corr/progressive} regard each individual frame as one sample and its nearby frames are fed into network together to provide temporal information for this frame. This method introduces redundant computation in adjacent frames since each frame is fed into the network as input for multiple times. In this paper, we seek to explore a more general and efficient temporal representation for GEBD task.

% The structured partition of sequence (SPoS) mechanism is employed to
%     re-partition input frame sequence and provides \textbf{structured context} for each candidate frame. Based on this structured context, the transformer encoder blocks are used to learn the high level representation of each local sequence, which have linear computational complexity with respect to the video length and enable feature sharing.

\noindent{\textbf{Structured Partition of Sequence.}} Given the video snippet $V = \{ I_t\}_{t=1}^T$, where $T$ is the time span of the video snippet and can be any arbitrary length, $I_t \in \mathbb{R}^C$ is the feature vector of frame $t$, which is generated from ResNet50 \cite{DBLP:conf/cvpr/resnet} backbone followed by a global average pooling layer, our goal is to obtain $K$ adjacent frames before candidate frame $I_t$ and $K$ adjacent frames after candidate frame $I_t$, where $K$ is the adjacent window size. We term this local sequence centred with candidate frame $I_t$ as \textbf{structured context} for frame $I_t$. To accomplish this while enabling feature sharing and maintaining efficiency and parallelism, we propose 
the novel Structured Partition of Sequence (SPoS) mechanism. Specifically, we first pad video $V = \{ I_t\}_{t=1}^T$ with $\operatorname{ceil}(\frac{T}{K}) \cdot K - T$ zero vectors at the end of the frame sequence so that the new video length $T'$ is divisible by $K$. Then given the padded video $V' \in \mathbb{R}^{T' \times C}$, we split it into $K$ slices where each slice $S_k$ ($k$ is the slice number, starts from $0$) is responsible to provide structured context frames for all $[k::K]$th frames (\ie, all frames that start from $k$ with a step of $K$). In this way, all video frames can be covered within all $K$ slices and these $K$ slices can be efficiently processed in parallel.

\begin{figure}[t]
    \centering
    \includegraphics[width=\textwidth]{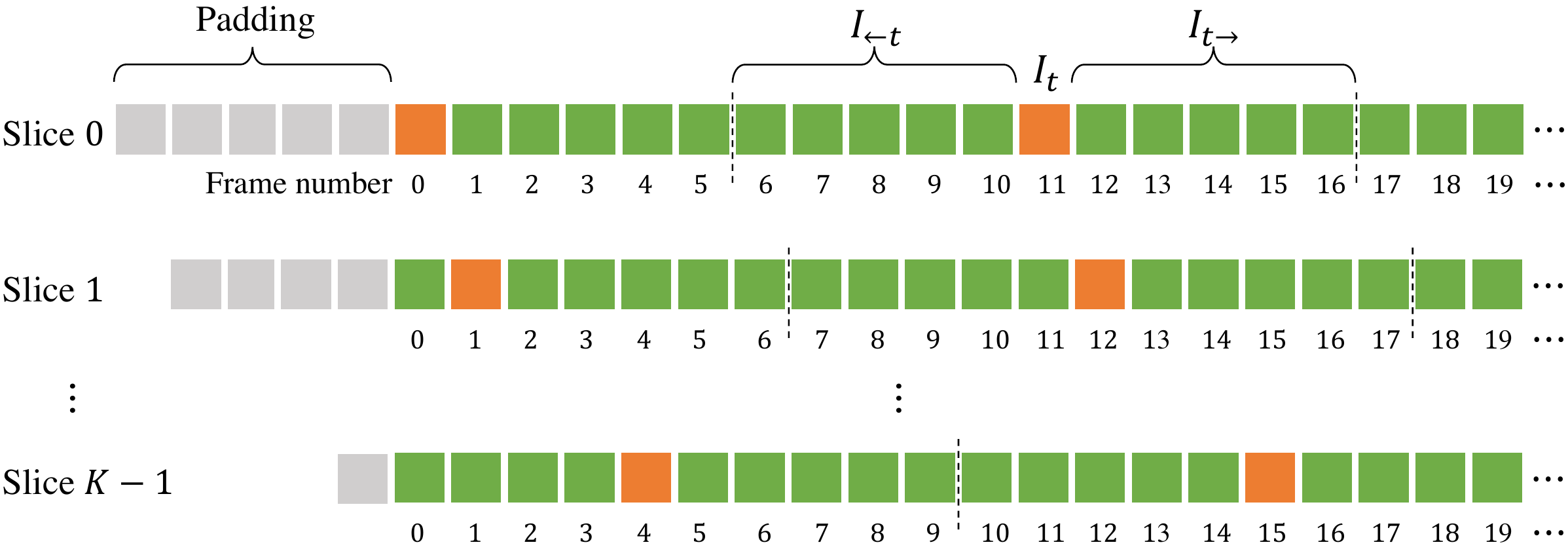}
    \caption{Illustration of proposed structured partition of sequence (SPoS). To obtain adjacent $K$ frames before candidate frame $I_t$ (denoted as $I_{\leftarrow t}$) and $K$ frames after $I_t$ (denoted as $I_{t \rightarrow}$), we split the input video sequence into $K$ slices. Each slice $S_k$ is responsible to produce adjacent frames $I_{\leftarrow t}$ and $I_{t \rightarrow}$ for the frames of specific indices (\ie, all frames that start from $k$ with a step of $K$). All video frames can be covered within all $K$ slices and can be efficiently processed in parallel. Our SPoS differs from Swin-Transformer \cite{DBLP:conf/iccv/swin-transformer} and 1D CNNs in that Swin-Transformer tends to learn a global representation after several stacks and is not structured and 1D CNNs actually make candidate frame $I_t$ attend to adjacent frames in a Gaussian distribution manner \cite{DBLP:conf/nips/LuoLUZ16}, respectively.}
    \label{fig:scr}
\end{figure}

In each frame slice $S_k$, we obtain structured context for frame $I_t$ in two directions, \ie, $K$ frames before frame $I_t$ and $K$ frames after frame $I_t$. We implement this through efficient memory view method provided by modern deep learning framework. Specifically, To obtain structured context frames $I_{\leftarrow t} \in \mathbb{R}^{K \times C}$ before frame $I_t$, we replicate the first frame of the padded video sequence $V'$ $K-k$ times and concatenate to the beginning of video $V'$ and the last $K-k$ frames of video $V'$ are dropped, and thus the number of frames is kept and still divisible by $K$. We denote this shifted video sequence as $V'_\leftarrow \in \mathbb{R}^{T' \times C}$. Then we view $V'_\leftarrow \in \mathbb{R}^{T' \times C}$ as $V_\leftarrow \in \mathbb{R}^{N \times K \times C}$, where $N = T' / K$ denotes the number of processed frames in slice $S_k$. In this way, we obtain the left structured context frames for all $N$ frames (\ie, all $[k::K]$th frames of origin video $V$). Similarly, to obtain structured context frames $I_{t\rightarrow} \in \mathbb{R}^{K \times C}$ after frame $I_t$, we replicate the last frame of the padded video sequence $k+1$ times and concatenate to the ending of video $V'$ and the first $k+1$ frames of video $V'$ are also dropped to keep the number of frames. In this way, we obtain the right structured context frames $V_\rightarrow \in \mathbb{R}^{N \times K \times C}$ for all $N$ frames. Finally, we can obtain all temporal context frames by repeating $K$ times for $K$ slices, and each frame $I_t$ is represented by its adjacent frames in a local window.

A key design element of our structured partition of sequence (SPoS) is its shared structured context information. We term this context information ``structured'' since SPoS maps each candidate frame $I_t$ to individual frame sequences $I_{\leftarrow t}$ and $I_{t\rightarrow}$ in a one-to-one manner, which is the key for accurate boundary detection. Our SPoS differs from Swin-Transformer \cite{DBLP:conf/iccv/swin-transformer} in that Swin-Transformer makes each frame able to attend very distant frames (\ie, tend to learn a global representation) due to its stacked shifted windows design. This is deleterious for boundary detection as very distant frames may cross multiple boundaries and thus provides less useful information. 
Another advantage of SPoS is that we can model these structured sequences using any sequential modeling method without considering computation complexity, due to its local shared and parallel nature, which can be computed in linear time to video length.

\noindent{\textbf{Encoding with Transformer.}} We use Transformer to model the structured context information. Given structured context features $I_{\leftarrow t} \in \mathbb{R}^{K \times C}, I_{t\rightarrow} \in \mathbb{R}^{K \times C}$of frame $I_t \in \mathbb{R}^C$, we first concatenate them in the temporal dimension to obtain context sequence $\mathbf{I}_t$ for frame $I_t$, \ie,
\begin{equation}
    \mathbf{I}_t = [I_{\leftarrow t}, I_t, I_{t\rightarrow}]
\end{equation}
where $\mathbf{I}_t \in \mathbb{R}^{L \times C}$, $L=2K+1$ and $[\cdot, \cdot, \cdot]$ denotes the concatenating operation. Then to model temporal information, we adapt a 6-layer Transformer \cite{DBLP:conf/nips/att_is_all_you_need} block to processing the context sequence $\mathbf{I}_t$ to get temporal representation $\mathbf{x}_t \in \mathbb{R}^{L \times C}$ within this structured context window. Unlike other methods \cite{DBLP:journals/corr/gebd-contrastive-learning,DBLP:journals/corr/gebd-UBoCo} where the computation of multi-head self attention (MSA) is based on global video frames sequence, our MSA computation is based only on the local temporal window. The computational complexity of the former is quadratic to video length $T$, \ie, $4TC^2 + 2T^2C$, and the computational complexity of our method is linear when $K$ is fixed (set to 8 by default, \ie, $L=17$), \ie, $4TC^2 + 2L^2TC$. Global self-attention computation is generally unaffordable for a large video length $T$, while our local structured based self-attention is scalable.

\subsection{Group Similarity}
The event boundaries of the GEBD task could be located at the moments where the action changes (\eg, Run to Jump), the subject changes (\eg, a new person appears), or the environment changes (\eg, suddenly become bright), for example. We experimentally observed that the frames within an adjacent local window provide more cues for event boundary detection than distant frames. This is consistent with human's intuition since the change of visual content can be regarded as an event boundary only in a short time period. Based on this observation, we can model local temporal information naturally based on structured context features extracted in Section \ref{sec:sc_transformer}.

The Transformer block aims at discovering relationships between frames and giving high level representation of frames sequence. However, event boundaries emphasize the differences between adjacent frames and neural networks tend to take shortcuts during learning \cite{DBLP:journals/natmi/shortcut-learning}. Thus classifying these frames directly into boundaries may lead to inferior performance due to non-explicit cues. Based on this intuition, we propose to guide classification with feature similarity of each frame pair in the structured temporal window $\mathbf{x}_t \in \mathbb{R}^{L \times C}$. Instead of performing similarity calculation with all $C$-dimensional channels, we found it beneficial to split the channels into several groups and calculate the similarity of each group independently. The concept of groups as a dimension for model
design has been more widely studied, including Group Convolutions \cite{DBLP:conf/nips/alexnet,DBLP:conf/cvpr/ResNeXt}, Group Normalization \cite{DBLP:conf/eccv/GN}, Multi-Head Self Attention \cite{DBLP:conf/nips/att_is_all_you_need}, \textit{etc}. However, to the best of our knowledge, there is still no study on similarity learning with grouping. Formally, given $\mathbf{x}_t \in \mathbb{R}^{L \times C}$, we first split into G groups:
\begin{equation}
    \mathbf{x'}_t = \texttt{reshape}(\mathbf{x}_t)
\end{equation}
where $\mathbf{x'}_t \in \mathbb{R}^{L \times G \times C'}$ and $C' = C / G$. Then the group similarity map $\mathbf{S}_t$ is calculated using the grouped feature:
\begin{equation}
\label{equa:sim_func}
    \mathbf{S}_t = \texttt{similarity-function}(\mathbf{x'}_t, \mathbf{x'}_t)
\end{equation}

\begin{figure}[t]
\centering
\begin{subfigure}{.24\textwidth}
    \centering
    \includegraphics[width=.99\linewidth]{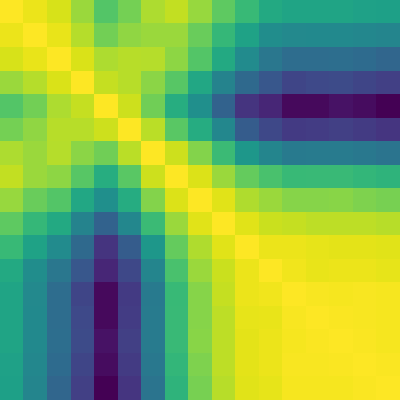}  
    % \caption{}
\end{subfigure}
\begin{subfigure}{.24\textwidth}
    \centering
    \includegraphics[width=.99\linewidth]{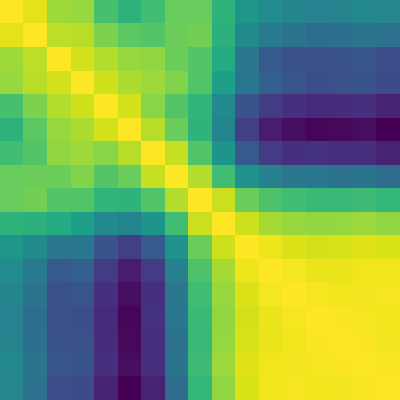}  
    % \caption{}
\end{subfigure}
\begin{subfigure}{.24\textwidth}
    \centering
    \includegraphics[width=.99\linewidth]{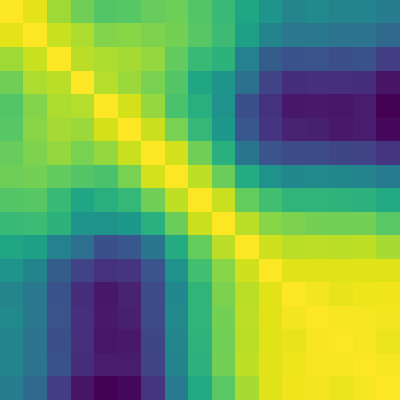}  
    % \caption{}
\end{subfigure}
\begin{subfigure}{.24\textwidth}
    \centering
    \includegraphics[width=.99\linewidth]{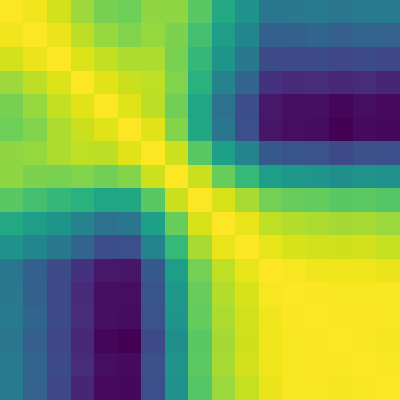}  
    % \caption{}
\end{subfigure}
\begin{subfigure}{.24\textwidth}
    \centering
    \includegraphics[width=.99\linewidth]{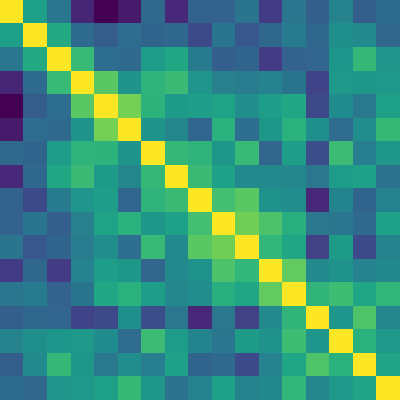}  
    % \caption{}
\end{subfigure}
\begin{subfigure}{.24\textwidth}
    \centering
    \includegraphics[width=.99\linewidth]{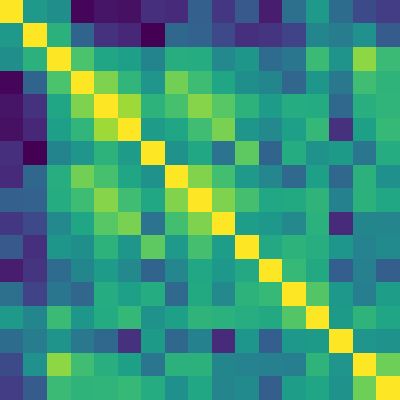}  
    % \caption{}
\end{subfigure}
\begin{subfigure}{.24\textwidth}
    \centering
    \includegraphics[width=.99\linewidth]{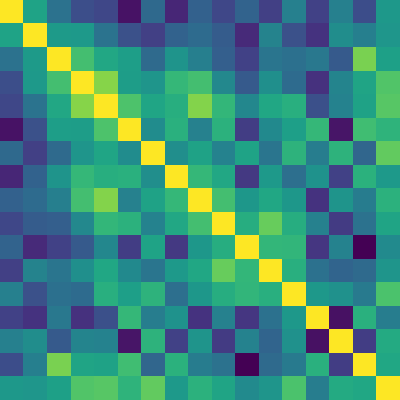}  
    % \caption{}
\end{subfigure}
\begin{subfigure}{.24\textwidth}
    \centering
    \includegraphics[width=.99\linewidth]{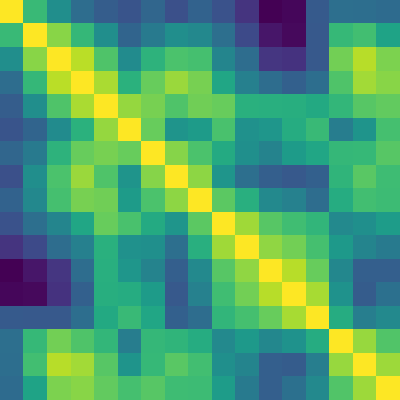}  
    % \caption{}
\end{subfigure}
\caption{Visualization of grouped similarity maps $\mathbf{S}_t$, $G=4$ in this example. First row indicates that there is a potential boundary in this local sequence while the second row shows no boundary in this sequence. We can also observe slightly different patterns between the same group, which may imply that each group is learning in a different aspect.}
\label{fig:similarity_maps}
\end{figure}

where $\mathbf{S}_t \in \mathbb{R}^{G \times L \times L}$, and $\texttt{similarity-function}(\cdot, \cdot)$ can be cosine similarity or euclidean similarity. As the group similarity map $\mathbf{S}_t$ contains the similarity patterns (efficient score of each frame pair, \ie, high response value when two frames are visually similar), it shows different patterns (as shown in Figure \ref{fig:similarity_maps}) in different sequences, which are critical for boundary detection. To keep our model as simple as possible, we use a 4-layer fully convolutional network \cite{DBLP:conf/cvpr/FCN} to learn the similarity patterns, which we found it work very well and efficient enough. Then we average pool the output of FCN to get a vector representation $h_t$, and this vector is used for downstream classification:
\begin{align}
\begin{array}{c}
    s_t = \operatorname{FCN}(\mathbf{S}_t) \\ 
    h_t = \texttt{average-pool}(s_t)
  \end{array}
\end{align}
where $s_t \in \mathbb{R}^{C \times L \times L}$ indicates the intermediate representation, $h_t \in \mathbb{R}^C$. The design principle of this module is extremely simple: computing group similarity patterns within local structured context based on previously encoded and using a small FCN to analyse the patterns.

\subsection{Optimization}
\label{sec:optim}
Our SC-Transformer and group similarity module are fully end-to-end, lightweight and in-place \textit{i.e.} no dimension change between input
and output. Therefore they can be directly used for further classification which is straightforward to implement and optimize. The video frame sequence $V = \{ I_t\}_{t=1}^T$ is represented by $\mathbf{V} = \{h_t\}_{t=1}^T$ after group similarity module, \ie, $\mathbf{V} \in \mathbb{R}^{T \times C}$. Then we stack 3 layers of 1D convolutional neural network to predict boundary scores. We use a single binary cross entropy loss to optimize our network.

GEBD is a taxonomy-free task and interconnects the mechanism of human perception to deep video understanding. The event boundary labels of each video are annotated by around 5 different annotators to capture human perception differences and therefore ensure diversity. However, this brings ambiguity of annotations and is hard for network to optimize, which may lead to poor convergence. To solve this issue and prevent the model from predicting the event boundaries too confidently, we use the Gaussion distribution to smooth the ground-truth boundary labels and obtain the soft labels instead of using the ``hard labels'' of boundaries. Specifically, for each annotated boundary, the intermediate label of the neighboring position $t'$ is computed as:
\begin{equation}
    \mathcal{L}_{t'}^t = \exp\Big( -\frac{( t-t' )^2}{2\sigma^2} \Big)
\end{equation}
where $\mathcal{L}_{t'}^t$ indicates the intermediate label at time $t'$ corresponding to the annotated boundaries at time $t$. We set $\sigma =1$ in all our experiments. The final soft labels are computed as the summation of all intermediate labels. Finally, binary cross entropy loss is used to minimize the difference between model predictions and the soft labels.

\section{Experiments}
We show that our method achieves competitive results compared to previous methods in quantitative evaluation on Kinetics-GEBD \cite{DBLP:journals/corr/GEBD} and TAPOS \cite{DBLP:conf/cvpr/TAPOS}. Then, we provide a detailed ablation study of different model design with insights and quantitative results.

\noindent{\textbf{Dataset.}} We perform experiments on both Kinetics-GEBD dataset \cite{DBLP:journals/corr/GEBD} and TAPOS dataset \cite{DBLP:conf/cvpr/TAPOS}. Kinetics-GEBD dataset consists of $54,691$ videos and $1,290,000$ temporal boundaries, which spans a broad spectrum of video domains in the wild and is open-vocabulary, taxonomy-free. Videos in Kinetics-GEBD dataset are randomly selected from Kinetics-400 \cite{DBLP:journals/corr/Kinetics}. The ratio of training, validation and testing videos of Kinetics-GEBD is nearly 1:1:1. Since the ground truth labels for the testing videos is not released, we train our model on training set and test on validation set. TAPOS dataset containing Olympics sport videos cross 21 action classes. The training set contains $13,094$ action instances and the validation set contains $1,790$ action instances. Following \cite{DBLP:journals/corr/GEBD}, we re-purpose TAPOS for GEBD task by trimming each action instance with its action label hidden and conducting experiments on each action instance. 

\noindent{\textbf{Evaluation Protocol.}}
To quantitatively evaluate the results of generic event boundary detection task, F1 score is used as the measurement metric. As described in \cite{DBLP:journals/corr/GEBD}, Rel.Dis. (Relative Distance, the error between the detected and ground truth timestamps, divided by the length of the corresponding whole action instance) is used to determine whether a detection is correct (\ie, $\leq$ threshold) or incorrect (\ie, $>$ threshold). A detection result is compared against each rater’s annotation and the highest F1 score is treated as the final result. We report F1 scores of different thresholds range from 0.05 to 0.5 with a step of 0.05.

\noindent{\textbf{Implementation Details.}}
For fair comparison with other methods, a ResNet50 \cite{DBLP:conf/cvpr/resnet} pretrained on ImageNet \cite{DBLP:conf/cvpr/imagenet} is used as the basic feature extractor in all experiments if not particularly indicated, note that we don't freeze the parameters of ResNet50 and they are optimized through backpropagation. Images are resized to 224$\times$224 following \cite{DBLP:journals/corr/GEBD}. We uniformly sample 100 frames from each video for batching purpose, \ie, $T=100$ in section \ref{sec:method}. We use the standard SGD with momentum set to $0.9$, weight decay set to $10^{-4}$, and learning rate set to $10^{-2}$. We set the batch size to $4$ (4 videos, equivalent to 400 frames) for each GPU and train the network on $8$ NVIDIA Tesla V100 GPUs, resulting in a total batch size of $32$, and automatic mixed precision training is used to reduce the memory burden. The network is trained for $30$ epochs with a learning rate drop by a factor of $10$ after $16$ epochs and $24$ epochs, respectively. All the source code of our method will be made publicly available after our paper is accepted.

\begin{figure}[t]
    \centering
    \includegraphics[width=\textwidth]{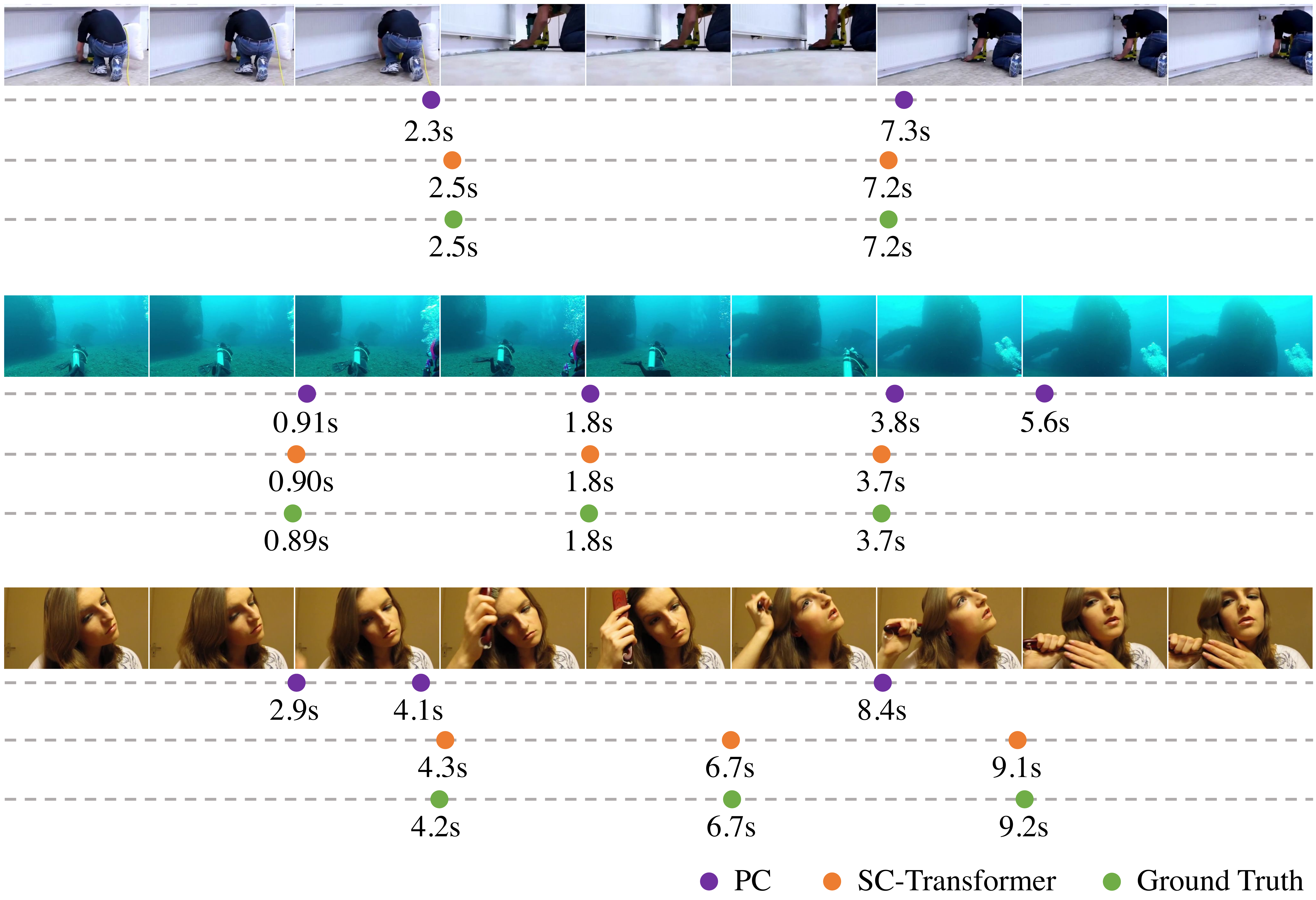}
    \caption{Example qualitative results on Kinetics-GEBD validation split. Compared with PC~\cite{DBLP:journals/corr/GEBD}, our SC-Transformer can generate more accurate boundaries which are consistent with ground truth.}
    \label{fig:visualization}
\end{figure}

% dim=512
%              0.05     0.1    0.15     0.2    0.25     0.3    0.35     0.4    0.45     0.5     Avg
% ---------  ------  ------  ------  ------  ------  ------  ------  ------  ------  ------  ------
% Recall     0.7905  0.8628  0.8856  0.8993  0.9082  0.9141  0.9182  0.9211  0.9238  0.9262  0.8950
% Precision  0.7586  0.8342  0.8577  0.8699  0.8767  0.8810  0.8840  0.8865  0.8889  0.8910  0.8629
% F1         0.7742  0.8483  0.8714  0.8844  0.8922  0.8973  0.9008  0.9035  0.9060  0.9082  0.8786

% dim=256
%              0.05     0.1    0.15     0.2    0.25     0.3    0.35     0.4    0.45     0.5     Avg
% ---------  ------  ------  ------  ------  ------  ------  ------  ------  ------  ------  ------
% Recall     0.7888  0.8596  0.8841  0.8984  0.9080  0.9137  0.9181  0.9214  0.9239  0.9263  0.8942
% Precision  0.7645  0.8389  0.8612  0.8743  0.8816  0.8863  0.8896  0.8921  0.8944  0.8966  0.8680
% F1         0.7765  0.8491  0.8725  0.8862  0.8946  0.8998  0.9036  0.9065  0.9089  0.9112  0.8809
% dim=128
%              0.05     0.1    0.15     0.2    0.25     0.3    0.35     0.4    0.45     0.5     Avg
% ---------  ------  ------  ------  ------  ------  ------  ------  ------  ------  ------  ------
% Recall     0.7741  0.8466  0.8708  0.8870  0.8969  0.9033  0.9080  0.9122  0.9151  0.9178  0.8832
% Precision  0.7766  0.8490  0.8723  0.8857  0.8935  0.8984  0.9016  0.9041  0.9062  0.9086  0.8796
% F1         0.7753  0.8478  0.8716  0.8864  0.8952  0.9009  0.9048  0.9081  0.9106  0.9132  0.8814

\begin{table}[t]
\centering
\caption{F1 results on Kinetics-GEBD validation split with Rel.Dis. threshold set from 0.05 to 0.5 with 0.05 interval.}
\scalebox{0.902}{
\begin{tabular}{lcccccccccc|c}
\toprule
Rel.Dis. threshold& 0.05 & 0.1 & 0.15 & 0.2 & 0.25 & 0.3 & 0.35 & 0.4 & 0.45 & 0.5 & avg \\ 
\midrule
BMN~\cite{DBLP:conf/iccv/BMN}& 0.186  & 0.204  & 0.213  & 0.220  & 0.226& 0.230& 0.233& 0.237& 0.239& 0.241& 0.223\\
BMN-StartEnd~\cite{DBLP:conf/iccv/BMN} & 0.491& 0.589& 0.627& 0.648& 0.660& 0.668& 0.674& 0.678& 0.681& 0.683& 0.640\\
TCN-TAPOS~\cite{DBLP:conf/eccv/TCN} & 0.464& 0.560& 0.602& 0.628& 0.645& 0.659& 0.669& 0.676& 0.682& 0.687& 0.627\\
TCN~\cite{DBLP:conf/eccv/TCN} & 0.588& 0.657& 0.679& 0.691& 0.698& 0.703& 0.706& 0.708& 0.710& 0.712& 0.685\\
PC~\cite{DBLP:journals/corr/GEBD}& 0.625& 0.758&0.804&0.829&0.844&0.853&0.859&0.864&0.867&0.870&0.817\\
SBoCo-Res50~\cite{DBLP:journals/corr/gebd-UBoCo} &0.732&-&-&-&-&-&-&-&-&-&0.866 \\
DDM-Net~\cite{DBLP:journals/corr/progressive}&0.764&0.843&0.866&0.880&0.887&0.892&0.895&0.898&0.900&0.902&0.873\\
\midrule
Ours &\textbf{0.777}&\textbf{0.849}&\textbf{0.873}&\textbf{0.886}&\textbf{0.895}&\textbf{0.900}&\textbf{0.904}&\textbf{0.907}&\textbf{0.909}&\textbf{0.911}&\textbf{0.881}\\
\bottomrule
\end{tabular}
}
\label{tab:gebd_val}
\vspace*{-0.1in}
\end{table}

\begin{table}[t]
\centering
\caption{F1 results on TAPOS validation split with Rel.Dis. threshold set from 0.05 to 0.5 with 0.05 interval.}
\scalebox{0.915}{
\begin{tabular}{lcccccccccc|c}
\toprule
Rel.Dis. threshold& 0.05 & 0.1 & 0.15 & 0.2 & 0.25 & 0.3 & 0.35 & 0.4 & 0.45 & 0.5 & avg \\ 
\midrule
ISBA~\cite{DBLP:conf/cvpr/ISBA} &0.106&0.170&0.227&0.265&0.298&0.326&0.348&0.369&0.382&0.396&0.302\\
TCN~\cite{DBLP:conf/eccv/TCN} & 0.237 & 0.312 & 0.331 & 0.339 & 0.342 & 0.344 & 0.347 & 0.348 & 0.348 & 0.348 & 0.330\\
CTM~\cite{DBLP:conf/eccv/CTM} & 0.244 & 0.312 & 0.336 & 0.351 & 0.361 & 0.369 & 0.374 & 0.381 & 0.383 & 0.385 & 0.350\\
TransParser~\cite{DBLP:conf/cvpr/TAPOS}& 0.289 & 0.381 & 0.435 & 0.475 & 0.500 & 0.514 & 0.527 & 0.534 & 0.540&0.545&0.474 \\
PC~\cite{DBLP:journals/corr/GEBD}& 0.522 & 0.595 & 0.628 & 0.646 & 0.659 & 0.665 & 0.671 & 0.676 & 0.679 & 0.683 & 0.642 \\
DDM-Net~\cite{DBLP:journals/corr/progressive}& 0.604 & 0.681 & 0.715 & 0.735 & 0.747 & 0.753 & 0.757 & 0.760 & 0.763 & 0.767 & 0.728 \\
\midrule
Ours &\textbf{0.618} &\textbf{0.694} &\textbf{0.728} &\textbf{0.749} &\textbf{0.761} &\textbf{0.767} &\textbf{0.771} &\textbf{0.774} &\textbf{0.777} &\textbf{0.780}& \textbf{0.742}\\
\bottomrule
\end{tabular}
}
\label{tab:tapos_val}
\end{table}

\subsection{Main Results}
\noindent\textbf{Kinetics-GEBD.} Table \ref{tab:gebd_val} illustrates the results of our models on Kinetics-GEBD validation set. Our method surpasses all previous methods in all Rel.Dis. threshold settings, demonstrating the effectiveness of structured partition of sequence and group similarity. Compared to the method PC~\cite{DBLP:journals/corr/GEBD}, our method achieves 15.2\% absolute improvement with $5.7\times$ faster running speed (\ie, 10.8ms per frame vs 1.9ms per frame). Compared to DDM-Net \cite{DBLP:journals/corr/progressive}, we also achieve 1.3\% absolute improvement. Since DDM-Net is not open sourced yet we are not able to compare runtime speed with our method. However it is worth noting that DDM-Net leverage the same input representation as PC~\cite{DBLP:journals/corr/GEBD}, \ie, each frame and it's adjacent frames are fed into network individually, which introducing many redundant computations. For example, given a video clip of length 100 and the window is set to 11 as mentioned in their paper, they have to process $1,100$ frames individually to get all boundary predictions for this single video. Our method is \texttt{independent} of video length and can get all boundary predictions in a single forward pass by just feeding the necessary 100 frames. The example qualitative results on Kinetics-GEBD are shown in Figure \ref{fig:visualization}.

\noindent\textbf{TAPOS.} We also conduct experiments on TAPOS dataset~\cite{DBLP:conf/cvpr/TAPOS}. TAPOS dataset contains Olympics sport videos with 21 actions and is not suitable for GEBD task. Following \cite{DBLP:journals/corr/GEBD}, we re-purpose TAPOS for GEBD task by trimming each action instance with its action label hidden, resulting in a more fine-grained sub-action boundary detection dataset. The results are presented in Table \ref{tab:tapos_val}. We boost F1@0.05 score by 9.6\% and 1.4\% compared with PC~\cite{DBLP:journals/corr/GEBD} and DDM-Net~\cite{DBLP:journals/corr/progressive}, respectively. This verified the effectiveness of our method and our method can learn more robust feature presentation in different scenes.

\subsection{Ablations}
Structured partition of sequence re-partition the video frame sequence into a more suitable format for GEBD task. Based on this unified and shared representation, we use simple yet effective group similarity to capture differences between frames. In our ablation analysis, we explore how each component of our method and loss influences the final performance. For the study we conduct experiments on Kinetics-GEBD dataset and use ResNet-50 as the backbone. In these experiments, we only present F1 score with $0.05$, $0.25$ and $0.5$ Rel.Dis. threshold due to limited space. \textit{Average} column indicates average F1 score of Rel.Dis. threshold set from 0.05 to 0.5 with 0.05 interval.

\noindent\textbf{Importance of structured partition of sequence (SPoS).}
Structured partition of sequence provides shared local temporal context for each frame to predict event boundaries. To verify its effectiveness, we attempt to remove it completely and use 1D convolution neural network and shifted window (Swin) representation \cite{DBLP:conf/iccv/swin-transformer} as replacements, results can be found in Table \ref{tab:ablation_window_representation}. We observed a significant performance drop after replacing SPoS. It can be interpreted that 1D CNNs only enlarge the receptive field of each candidate frame and this impact actually distributes as a Gaussian \cite{DBLP:conf/nips/LuoLUZ16}. This is not optimal for event boundary detection since nearby frames may have equal importance. As for Swin~\cite{DBLP:conf/iccv/swin-transformer}, it's designed to relieve Transformer's global self-attention computation burden by leveraging non-overlapped shifted windows. And each frame can attend to very distant frames after several Swin Transformer Block stacks. We think this is \texttt{not aligned} with GEBD task since adjacent frames are more important while distant frames may cross multiple different boundaries and thus disturb the convergence. This also verifies that structured representation is crucial for accurate boundary detection.

\begin{table}[t]
    \centering
    \caption{Importance of structured partition of sequence (SPoS). When replacing our SPoS with 1D convolution neural network and Swin-Transformer~\cite{DBLP:conf/iccv/swin-transformer} (non-overlapping shifted window representation), we can observe a significant performance drop. $\Delta$ rows show the differences with our SPoS. This verifies that SPoS is crucial for boundary detection.}
    \setlength{\tabcolsep}{4pt}
    \begin{tabular}{l|cc|cc|cc|cc}
    \toprule
         Representation  & 0.05 &$\Delta$  &0.25 &$\Delta$& 0.5&$\Delta$ & Average &$\Delta$ \\
    \midrule
         1D CNN&0.609&-0.168 &0.838&-0.057 &0.864&-0.044 &0.810&-0.071  \\
         Swin~\cite{DBLP:conf/iccv/swin-transformer}&0.703&-0.074 &0.870&-0.025 &0.891&-0.017 & 0.849&-0.032  \\
         SPoS&\textbf{0.777}&- &\textbf{0.895}&- &\textbf{0.911}&- &\textbf{0.881}&- \\
    \bottomrule
    \end{tabular}
    \vspace*{-0.1in}
    \label{tab:ablation_window_representation}
\end{table}

\noindent\textbf{Adjacent window size $K$.} Adjacent window size $K$ defines how far can the subsequent module capture context information in the temporal domain. A smaller $K$ may not be able to capture enough necessary context information for a boundary while a larger $K$ will introduce noise information when cross two or more different boundaries. As presented in Table \ref{tab:ablation_k}, we observed different F1 scores after varying $K$. We believe that event boundaries in a video may span different number of frames to recognize them. Hence intuitively, different kinds of boundaries may prefer to different window size $K$. Although more sophisticated mechanism like adapting $K$ size may further increase the performance, we choose a fixed-length window in all our experiments for simplicity and remain this as a future work. The performance gain diminishes as $K$ increases, and we choose $K=8$ as the adjacent window size.

\begin{table}[t]
    \centering
    \caption{Effect of adjacent window size $K$. Different F1 scores are observed after varying $K$. This can be interpreted as that a smaller $K$ may not be able to capture enough necessary context information for a boundary while a larger $K$ will introduce noise information when cross two or more different boundaries.}
    \setlength{\tabcolsep}{8pt}
    \begin{tabular}{c|cccc}
    \toprule
        Window size $K$  & 0.05 &0.25 & 0.5 & Average \\
    \midrule
         $2$&0.745 &0.854 &0.869 & 0.842 \\
         $4$&0.762 &0.881 &0.897 & 0.867  \\
         $6$&0.771 &0.889 &0.904 & 0.875 \\
         $8$&\textbf{0.777} &0.895 &0.911 &0.881  \\
         $10$&0.776 &0.894 &\textbf{0.912} &0.880  \\
         $12$&0.777 &\textbf{0.896} &\textbf{0.912} &\textbf{0.882}  \\
    \bottomrule
    \end{tabular}
    \vspace*{-0.1in}
    \label{tab:ablation_k}
\end{table}

\noindent\textbf{Effect of model width.} In Table \ref{tab:ablation_channel} we study the model width (number of channels). We use $C=256$ by default and it has the best performance.

\begin{table}[t]
        \begin{minipage}{0.5\textwidth}
            \centering
            \caption{Effect of model width $C$.}
            \setlength{\tabcolsep}{4pt}
            \begin{tabular}{c|cccc}
            \toprule
                 $C$  & 0.05 &0.25 & 0.5 & Average \\
            \midrule
                 $128$&0.775 &\textbf{0.895} &\textbf{0.913} &\textbf{0.881}  \\
                 $256$&\textbf{0.777} &\textbf{0.895} &0.911 & \textbf{0.881} \\
                 $512$&0.774 &0.892 &0.908 & 0.879  \\
                 $768$&0.768  &0.887 &0.904 &0.875  \\
                 $1024$&0.770 &0.889 &0.905 &0.876 \\
            \bottomrule
            \end{tabular}
            \label{tab:ablation_channel}
            \vspace*{-0.1in}
        \end{minipage}
        \hfillx
        \begin{minipage}{0.5\textwidth}
            \centering
            \caption{Effect of number of groups $G$.}
            \setlength{\tabcolsep}{4pt}
            \begin{tabular}{c|cccc}
            \toprule
                 $G$  & 0.05 &0.25 & 0.5 & Average \\
            \midrule
                 $1$&0.761 &0.871 &0.887 &0.861  \\
                 $2$&0.769 &0.891 &0.907 & 0.877 \\
                 $4$&0.777 &0.895 &0.911 & 0.881  \\
                 $8$&\textbf{0.778} &\textbf{0.896} &\textbf{0.913} &\textbf{0.882}  \\
                 $16$&0.777 &\textbf{0.896} &0.912 &0.881  \\
            \bottomrule
            \end{tabular}
            \label{tab:ablation_group}
            \vspace*{-0.1in}
        \end{minipage}
    \end{table}

% \begin{table}[ht]
%     \centering
%     \caption{Effect of model width $C$.}
%     \setlength{\tabcolsep}{5pt}
%     \begin{tabular}{c|cccc}
%     \toprule
%          $C$  & 0.05 &0.25 & 0.5 & Average \\
%     \midrule
%          $128$&0.775 &0.895 &0.913 &0.881  \\
%          $256$&0.777 &0.895 &0.911 & 0.881 \\
%          $512$&0.774 &0.892 &0.908 & 0.879  \\
%          $768$& & & &  \\
%          $1024$&0.776 &0.894 &0.911 &0.880 \\
%     \bottomrule
%     \end{tabular}
%     \label{tab:ablation_channel}
% \end{table}

\noindent\textbf{Number of groups.} We evaluate the importance of group similarity by changing the number of groups $G$, results are shown in Table \ref{tab:ablation_group}. We observe steady performance improvements when increasing $G$ and saturated when $G=4$. This result shows the effectiveness of grouping channels when computing similarity.

% \begin{table}[ht]
%     \centering
%     \caption{Effect of number of groups $G$.}
%     \setlength{\tabcolsep}{5pt}
%     \begin{tabular}{c|cccc}
%     \toprule
%          $G$  & 0.05 &0.25 & 0.5 & Average \\
%     \midrule
%          $1$&0.761 &0.871 &0.887 &0.861  \\
%          $2$&0.769 &0.891 &0.907 & 0.877 \\
%          $4$&0.774 &0.892 &0.908 & 0.879  \\
%          $8$&\textbf{0.775} &\textbf{0.893} &\textbf{0.910} &\textbf{0.880}  \\
%          $16$&0.774 &\textbf{0.893} &0.909 &0.879  \\
%     \bottomrule
%     \end{tabular}
%     \label{tab:ablation_group}
% \end{table}

\noindent\textbf{Effect of similarity function.} We explore different distance metrics (we call them similarity since minus value is used) in Table \ref{tab:ablation_sim_func}. The results show that our method is effective to different metrics, and we use cosine metric in our experiments.

\begin{table}[ht]
    \centering
    \caption{Effect of $\texttt{similarity-function}(\cdot, \cdot)$ in Equation \ref{equa:sim_func}.}
    \setlength{\tabcolsep}{5pt}
    \begin{tabular}{l|cccc}
    \toprule
         Function  & 0.05 &0.25 & 0.5 & Average \\
    \midrule
         Chebyshev&0.770 &0.887 &0.905 &0.872  \\
         Manhattan&0.774 &0.894 &0.907 &0.878  \\
         Euclidean&0.776 &\textbf{0.895} &0.910 &\textbf{0.881}  \\
         Cosine   &\textbf{0.777} &\textbf{0.895} &\textbf{0.911} &\textbf{0.881} \\
    \bottomrule
    \end{tabular}
    \vspace*{-0.15in}
    \label{tab:ablation_sim_func}
\end{table}

\noindent\textbf{Loss ablations.} GEBD task can be regarded as a framewise binary classification (boundary or not) after capturing temporal context information. We train our model with binary cross entropy (BCE) loss and mean squared error (MSE) loss with turning Gaussian smoothing (introduced in section \ref{sec:optim}) on and off. As shown in Table \ref{tab:ablation_loss}, Gaussian smoothing can improve the performance in both settings, which shows its effectiveness. We attribute this improvement to two aspects: 1) Consecutive frames have similar feature representation in the latent space thus consecutive frames are always tend to output closer responses, hard labels violate this rule and lead to poor convergence. 2) Annotations of GEBD 
are ambiguous in nature and Gaussian smoothing prevents the network from becoming overconfident. We use ``BCE + Gaussian'' setting in all our experiments.
\vspace*{-0.3in}
\begin{table}[ht]
    \centering
    \caption{Effect of loss function.}
    \setlength{\tabcolsep}{5pt}
    \begin{tabular}{ccc|cccc}
    \toprule
         BCE & MSE &Gaussian  & 0.05 &0.25 & 0.5 & Average \\
    \midrule
        &\checkmark&  &0.758 &0.881 &0.899 &0.865  \\
        &\checkmark&\checkmark  &0.771 &0.893 &0.910 &0.877  \\
        \checkmark &&&0.763 &0.887 &0.905 &0.872  \\
        \checkmark &&\checkmark &\textbf{0.777} &\textbf{0.895} &\textbf{0.911} &\textbf{ 0.881}  \\
    \bottomrule
    \end{tabular}
    \vspace*{-0.35in}
    \label{tab:ablation_loss}
\end{table}

\section{Conclusions}
In this work, we presented SC-Transformer which is a fully end-to-end method for generic event boundary detection. Structured partition of sequence mechanism is proposed to provide structured context information for GEBD task and Transformer encoder is adapted to learn high-level representation. Then group similarity and FCN are used to exploit discriminative features to make accurate predictions. Gaussian kernel is used to preprocess the ground-truth annotations to speed up training process. The proposed method achieves start-of-the-art results on the challenging Kinetics-GEBD and TAPOS datasets with high running speed. We hope our method can inspire future work.

% \clearpage
% \bibliography{eccv2022submission}

\end{document}